\documentclass[10pt,twocolumn,letterpaper]{article}

\usepackage[pagenumbers]{iccv} 

\usepackage{tabularx}
\usepackage{booktabs}
\usepackage[inkscapelatex=false]{svg}
\usepackage[nolist]{acronym}
\begin{acronym}
\acro{NAME}[VisualSpeaker]{VisualSpeaker}
\acro{3dgs}[3DGS]{3D Gaussian Splatting}
\acro{3dmm}[3DMM]{3D Morphable Model}
\acro{bsl}[BSL]{British Sign Language}

\acro{cnn}[CNN]{Convolutional Neural Network}
\acro{rnn}[RNN]{Recurrent Neural Network}
\acro{tcn}[TCN]{Temporal Convolutional Network}
\acro{vae}[VAE]{Variable Autoencoder}

\acro{dnn}[DNN]{Deep Neural Network}
\acro{nlp}[NLP]{Natural Language Processing}
\acro{facs}[FACS]{Facial Action Coding System}
\acro{fau}[FAU]{Facial Action Unit}
\acro{fid}[FID]{Fréchet Inception Distance}
\acro{flame}[FLAME]{Faces Learned with an Articulated Model and Expressions}

\acro{gan}[GAN]{Generative Adversarial Networks}
\acro{gpu}[GPU]{Graphics Processing Unit}

\acro{lsed}[LSE-D]{Lip Sync Error - Distance}
\acro{lsec}[LSE-C]{Lip Sync Error - Confidence}
\acro{lve}[LVE]{Lip Vertex Error}
\acro{lre}[LrE]{Lip Reading Error}

\acro{ldm}[LDM]{Latent Diffusion Model}
\acro{lmd}[LMD]{Landmark Distance}
\acro{lpips}[LPIPS]{Learned Perceptual Image Patch Similarity}

\acro{mae}[MAE]{Masked Autoencoders}
\acro{mlp}[MLP]{Multi-Layer Perceptron}
\acro{mse}[MSE]{Mean Squared Error}
\acro{mead}[MEAD]{Multi-view Emotional Audio-visual Dataset}

\acro{nerf}[NeRF]{Neural Radiance Fields}
\acro{nphm}[NPHM]{Neural Parametric Head Model}
\acro{nmf}[NMF]{Non-Manual Feature}
\acro{nvs}[NVS]{Novel View Synthesis}

\acro{pca}[PCA]{Principal Component Analysis}
\acro{psnr}[PSNR]{Peak Signal-to-Noise}

\acro{sdf}[SDF]{Signed Distance Field}
\acro{sfm}[SfM]{Structure from Motion}
\acro{sl}[SL]{Sign Language}
\acro{slp}[SLP]{Sign Language Production}

\acro{sota}[SOTA]{State-of-the-Art}
\acro{ssim}[SSIM]{Structural Similarity Index Measure}
\acro{tcn}[TCN]{Temporal Convolutional Network}
\acro{tts}[TTS]{Text-to-Speech}
\acro{vqvae}[VQ-VAE]{Vector Quantized-Variational AutoEncoder}
\acro{vasr}[V-ASR]{Visual Automatic Speech Recognition}
\end{acronym}
\usepackage{float}
\usepackage{afterpage}
\usepackage{subcaption}
\usepackage{multirow}
\usepackage{graphicx}
\usepackage{siunitx}
\usepackage{xcolor}

\definecolor{iccvblue}{rgb}{0.21,0.49,0.74}
\usepackage[pagebackref,breaklinks,colorlinks,allcolors=iccvblue]{hyperref}

\def\paperID{8} 
\def\confName{ICCV}
\def\confYear{2025}

\title{\acs{NAME}: Visually-Guided 3D Avatar Lip Synthesis}

\author{Alexandre Symeonidis-Herzig\hspace{0.6cm} {\"O}zge Mercano\u{g}lu Sincan\hspace{0.6cm} Richard Bowden \\[0.2cm]
CVSSP, University of Surrey, United Kingdom\\
\small{\texttt{\{a.symeonidisherzig, o.mercanoglusincan, r.bowden\}@surrey.ac.uk}}
}

\begin{document}

\twocolumn[{
\renewcommand\twocolumn[1][]{#1}
\maketitle
\begin{center}
    \captionsetup{type=figure}
    \includegraphics[width=\textwidth]{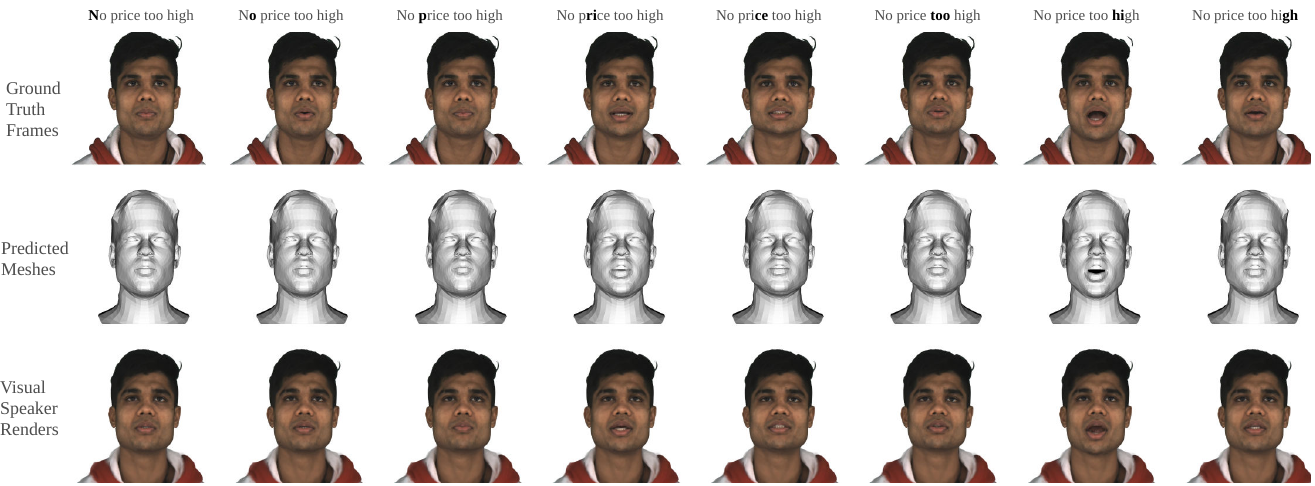}
    \captionof{figure}{
    \textbf{\acs{NAME} results.} The generated animation from the phrase “No price too high.” Ground truth video (top), the FLAME meshes predicted by our approach (middle), and the \acs{3dgs} renders driven by these meshes (bottom). Note how \acs{NAME} synthesizes lip movements that accurately and expressively articulate the input, achieved by combining geometric and perceptual supervision.}
    \label{fig:teaser}
\end{center}
}]

\begin{abstract}
Realistic, high-fidelity 3D facial animations are essential for expressive avatars in human-computer interaction and accessibility.  
Although prior methods show promising quality, their reliance on the mesh domain limits their ability to fully leverage the rapid visual innovations seen in 2D computer vision and graphics.
We propose \textbf{\acs{NAME}}, a novel method that bridges this gap using photorealistic differentiable rendering, supervised by visual speech recognition, for improved 3D facial animation.
Our contribution is a perceptual lip-reading loss, derived by passing photorealistic \acl{3dgs} avatar renders through a pre-trained \acl{vasr} model during training.
Evaluation on the MEAD dataset demonstrates that \acs{NAME} improves both the standard \acl{lve} metric by 56.1\% and the perceptual quality of the generated animations, while retaining the controllability of mesh-driven animation.
This perceptual focus naturally supports accurate mouthings, essential cues that disambiguate similar manual signs in sign language avatars.
\end{abstract}
    
\section{Introduction} \label{sec:intro}
People are innately skilled at recognizing and interpreting subtle facial cues, making the task of animating photorealistic 3D head avatars a challenging task as the renders face intense visual scrutiny.
Demand for such avatars is increasing in areas like telepresence, interactive media, and particularly \ac{slp}.
While spoken language uses mouth movements as secondary cues, sign language relies on mouthings as primary linguistic components to disambiguate similar manual signs \cite{Sutton-SpenceRachel2000TloB}.
Therefore, perceptually accurate mouthings are vital for effective sign language communication.

Despite this, prior 3D talking head methods \cite{faceformer2022,voca,xing2023codetalker,richard2021meshtalk} were predominantly guided by geometry, relying on \ac{mse} loss over the vertices to ensure the predicted mesh vertices closely match the ground truth.
While effective for reducing geometric error, these losses often yield averaged and minimally expressive animations that fail to differentiate visemes \cite{fisher1968confusions_visemes}, the basic visual units of speech that represent distinct mouth shapes and movements.
This occurs as geometric guidance alone is disconnected from how people perceive lip movements, especially when solely reading lips.
Vertex losses' inherent shortcomings have led to the explorations of additional losses, with Chae-Yeon \textit{et al.}~\cite{chaeyeon2025perceptuallyaccurate3dtalking} succinctly summarizing the three criteria for perceptually accurate lip movements as temporal synchronization, expressiveness, and lip readability.

Among these criteria, we focus on lip readability, which is critical in applications such as \ac{slp} and accessibility systems for deaf and hard-of-hearing users, where visemes act as linguistic signals that disambiguate manual signs and encode grammatical markers \cite{Sutton-SpenceRachel2000TloB}.
While recent approaches \cite{emote,eungi2024enhancingspeechdriven3dfacial} have targeted this using mesh-level losses, such geometric proxies remain fundamentally disconnected from the end goal: generating photorealistic and intelligible renders.
To bridge this gap between geometry and perception, we propose \acs{NAME}, a framework that directly optimizes for lip readability in the rendered pixel space.
Our method guides an autoregressive transformer by leveraging a pretrained \ac{vasr} model \cite{autoavsr_ma2023} to provide feedback on differentiable \ac{3dgs} renders \cite{kerbl20233dgaussiansplatting}.
This approach improves fidelity while enabling novel applications, such as generating silent mouthing animations for sign language directly from text.
Our main contributions are:
\begin{itemize}
\item A novel lip-reading loss computed directly on photorealistic \ac{3dgs} renders, closing the loop between geometric generation and perceptual evaluation.
\item Evaluations demonstrating that our method surpasses the mesh-based baselines in both geometric terms and, crucially, in terms of lip-readability, validated by a user study on intelligibility.
\item A text-to-mouthing application for \ac{slp}, which integrates a \ac{tts} system to generate accurate signing avatars from text alone, enabling scalable and audio-free avatar generation.
\end{itemize}

\section{Related Work}
\label{sec:related}
Animating speech-driven head avatars has been a long-standing research challenge, spanning decades of work across both 2D and 3D domains \cite{morishima98_avsp, Bregler_video_rewrite, zhang2023sadtalkerlearningrealistic3d, faceformer2022, voca, massaro1999picture_baldi}.
The field has progressed from rule-based systems to data-driven models, with growing emphasis on photorealism and perceptual quality.

\textbf{Linguistic Approaches.}
Rules-based procedural approaches to facial animation, often following the dominance model \cite{cohen1990synthesis}, remain prevalent in production environments, such as JALI \cite{edwards_jali} and FaceFX \cite{FaceFX2025}. 
These methods decompose speech into phonetic units, then apply hand-crafted mapping functions that transform these units into facial poses.
This yields consistent, controllable animation adopted in industry, but these methods require significant linguistic and artistic expertise to adapt to new languages or identities, limiting their scalability.

\textbf{Learning-based Approaches.}
To overcome the limitations of procedural systems, data-driven approaches learn animation directly from paired audio-visual data. 
Early methods \cite{morishima98_avsp,massaro1999picture_baldi} demonstrated feasibility but were limited by small datasets and computational constraints.
Subsequent methods based on RNNs and CNNs \cite{deeplearning_mesh_talking_snyth, voca, richard2021meshtalk, karras2017audio} enabled more expressive animations across diverse appearances.
However, even with attempts \cite{karras2017audio,richard2021meshtalk} to encapsulate the longer-term dependencies of speech, these methods still struggled to capture the complex relationships between audio and facial movements over a broad range of identities.
By adapting the Transformer architecture \cite{vaswani2023attentionneed}, models like FaceFormer \cite{faceformer2022} now better capture these relationships, achieving more expressive and temporally stable results by considering a longer audio context.

Despite these advances, over-smoothing remains a persistent challenge across all methods, leading to less expressive animations.
This primarily stems from the reliance on L2 losses in the geometric domain, which tend to average out subtle movements.
To mitigate this, works have proposed various strategies.
One approach, seen in CodeTalker \cite{xing2023codetalker}, uses a \acl{vqvae} codebook \cite{oord2018neuraldiscreterepresentationlearning} to discretize facial motions and preserve nuance through quantization.
Other methods move beyond simple MSE by incorporating auxiliary losses for lip-reading, emotion, or synchronization \cite{emote,zhuang2024learn2talk3dtalkingface,chaeyeon2025perceptuallyaccurate3dtalking}. While these losses improve geometric quality, they create a fundamental disconnect between the optimization target and the final, photorealistic visual output.

\textbf{Photorealistic Avatars.} Recent advances in novel view synthesis, particularly the real-time, differentiable rendering of 3D Gaussian Splatting (\ac{3dgs}) \cite{kerbl20233dgaussiansplatting}, have enabled the creation of high-fidelity avatars that can be animated in real-time.
This capability has opened up new approaches for 3D talking head synthesis \cite{guo2021adnerfaudiodrivenneural, li2024s3dnerfsingleshotspeechdrivenneural,Yu_2024_gaussiantalker,li2024talkinggaussians}, allowing direct animation of photorealistic 3D heads.
The rendering efficiency of \ac{3dgs} enables methods \cite{li2024talkinggaussians, Yu_2024_gaussiantalker} to incorporate visual losses directly during training.
For example, TalkingGaussian \cite{li2024talkinggaussians} creates two motion-fields, one for the head and one for the mouth, and uses an MLP conditioned on audio features to predict Gaussian primitives to render.
Training uses only visual losses on a few minutes of identity-specific data, yielding high-quality reconstructions but with limited generalization to unseen identities and no explicit control over other factors such as gaze or expressions.
GaussianTalker \cite{Yu_2024_gaussiantalker}, most related to our work, employs a mesh to drive the \ac{3dgs} avatar.
However, it relies on mesh-based renders for computing losses in a latent lip-reading space and photometric supervision of the rendered avatars, rather than directly evaluating lip readability in the final output.
This highlights a core limitation of mesh-based pipelines: when the mesh serves only as an intermediate geometry proxy, it cannot guarantee that fine-scale lip details are preserved after neural rendering.
Subtle articulatory cues such as tongue position, lip closure, and inner-mouth geometry can be smoothed out or misaligned if supervision stops at the mesh stage.
This disconnect between perceptual lip accuracy and training objectives motivates the need for supervision that directly enforces lip intelligibility in the final photorealistic output, as perceived by a human observer.

\section{Methodology}

\acs{NAME} combines parametric 3D face modeling, differentiable photorealistic rendering, and visual speech recognition to improve lip-synchronized facial animation.
This section first outlines the key components: the FLAME head model, 3D Gaussian Splatting avatars, the audio-text feature extractor, and the perceptual supervision module.
Then we detail the full architecture and training procedure.

\subsection{Preliminaries} \label{sec:prelim}
FLAME \cite{FLAME} is a widely adopted parametric 3D face model that represents head geometry using a compact set of interpretable parameters for identity ($\beta \in \mathbb{R}^{300}$), expression ($\psi \in \mathbb{R}^{100}$), and pose ($\theta \in \mathbb{R}^{6}$).
As a statistical \ac{3dmm} \cite{blanzvetter3dmm}, it deforms a canonical mesh via linear blend skinning:

\begin{equation}
F(\beta,\theta,\psi)\rightarrow(\textbf{V},\textbf{F}).
\label{eq:flame}
\end{equation}
Here, $\textbf{V} \in \mathbb{R}^{5142\times3}$ and $\textbf{F} \in \mathbb{Z}^{10144\times3}$ are the vertices and faces of the FLAME mesh, respectively.  

We adopt FLAME for its consistent topology and explicit factorization of static identity from dynamic expression, enabling robust mesh fitting and controllable animation.
As a de facto standard, FLAME has been extended in numerous works \cite{SMPL-X:2019, sklyarova2023neuralhaircutpriorguidedstrandbased_flamehair, Papantoniou_2022_CVPR_NED}, most relevant are works \cite{qian2024gaussianavatars} that allow for rendering using \ac{3dgs} \cite{kerbl20233dgaussiansplatting}. 
To accommodate the rendering, we rigidly attached 120 triangles representing teeth to the standard FLAME topology.

For datasets lacking 3D ground truth, we employ optimization-based tracking to obtain FLAME parameters for a 3D pseudo-ground truth.
Following VHAP \cite{qian2024versatile}, we fit FLAME to multi-view images from MEAD \cite{kaisiyuan2020mead} through a multi-stage approach.
Initial stages align the mesh using landmark-based losses, while later stages incorporate photometric losses on differentiable mesh renders \cite{ravi2020accelerating3ddeeplearning}.
We add temporal regularization to minimize jitter between frames and extend VHAP to refine camera parameters initially estimated via structure-from-motion \cite{colmap_sfm}.
For training, we also remove all head translation and rotation, as we focus on the facial expressions.
This pseudo-ground truth, while robust, has limitations.
The optimization-based fitting can sometimes struggle with extreme or very rapid expressions, and its accuracy is sensitive to the initial landmark detection.
This results in a 'noisier' ground truth compared to direct 3D scans , which motivates our curriculum learning strategy and the adjusted vertex weighting in later training stages.

\textbf{3DGS Avatar}. 
To generate photorealistic 3D head avatars, we employ \ac{3dgs} \cite{kerbl20233dgaussiansplatting}, which models appearance as a set of 3D Gaussian primitives with explicit positions, shapes, and view-dependent spherical harmonics. 
Unlike implicit \acs{nerf}, which typically have high computational costs for rendering, \ac{3dgs} supports differentiable, high-fidelity rendering at interactive rates.
This efficiency is critical for our method, making it computationally feasible to incorporate the perceptual lip-reading loss directly into the training loop without prohibitive overhead.

We bind the \ac{3dgs} primitives to a FLAME mesh following \cite{qian2024gaussianavatars}, allowing photorealistic rendering controlled with mesh deformations.
The fitting is optimization based and results in per-subject avatars, $G$.
\ac{3dgs}'s explicit nature enables differentiable rendering at interactive speeds.
Optimization produces accurate results, but is slow and depends on varied input views and sequences.
If these limitations were a concern, feed-forward approaches \cite{zheng2024headgapfewshot3dhead, zhang2025guavageneralizableupperbody, kirschstein2025avat3rlargeanimatablegaussian} could be used at minimal cost to final output quality.

During training, we render frames by passing the predicted FLAME mesh $(\textbf{V},\textbf{F})$, precomputed Gaussian parameters $G$, and camera parameters $C$ to the differentiable \ac{3dgs} renderer \cite{kerbl20233dgaussiansplatting} producing an image $\textbf{I} \in \mathbb{R}^{96\times96\times3}$:

\begin{equation}
    R(\textbf{V},\textbf{F},G,C)\rightarrow\textbf{I}
    \label{eq:render}
\end{equation}

\subsection{Perceptual Supervision}
\begin{figure}[b]
    \begin{subfigure}{0.32\linewidth}
        \includegraphics[width=\linewidth]{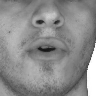}
        \caption{Ground Truth}
    \end{subfigure}
    \hfill
    \begin{subfigure}{0.32\linewidth}
        \includegraphics[width=\linewidth]{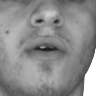}
        \caption{3DGS Render}
    \end{subfigure}
    \hfill
    \begin{subfigure}{0.32\linewidth}
        \includegraphics[width=\linewidth]{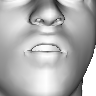}
        \caption{Mesh Render}
    \end{subfigure}
    \caption{\textbf{Lip Region Comparison.} Visual comparison of lip regions after alignment, cropping, and grayscale conversion for lip-reading supervision. The \ac{3dgs} render (middle) closely resembles the ground truth (left), while the mesh render (right) lacks photorealistic detail.}
    \label{fig:lips}
\end{figure}

\begin{figure*}
    \includegraphics[width=.95\linewidth]{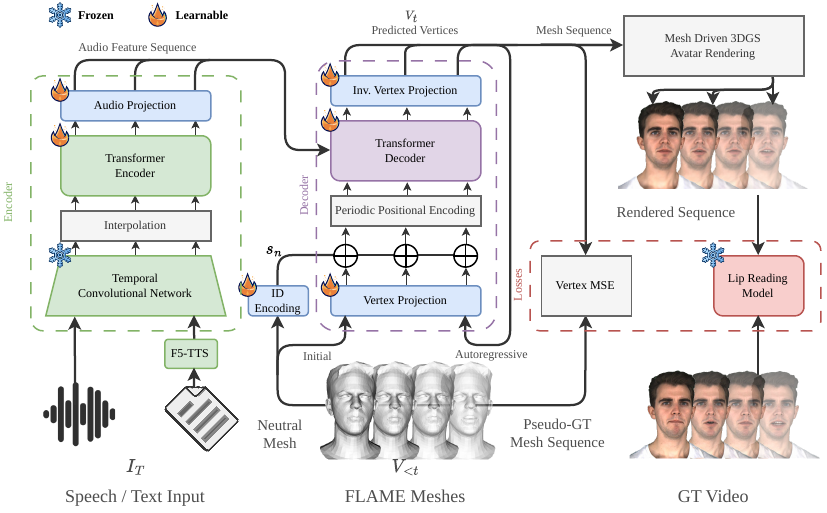}
    \caption{\textbf{Overview of \acs{NAME}.} Our encoder–decoder framework predicts the next frame’s vertex offsets, $\textbf{V}_t$. Given either text or audio, the encoder (left) generates input features $I_T$. These, together with past facial motion, $\textbf{V}_{<t}$, and a speaker identity embedding, $s_n$, derived from a neutral FLAME mesh, are passed to the decoder (middle). During training (right), predictions are supervised by a standard vertex loss and a novel perceptual loss computed on photorealistic 3DGS renders using a pretrained lip-reading model~\cite{autoavsr_ma2023}.}
    \label{fig:arch}
\end{figure*}

We introduce a novel perceptual supervision signal that evaluates lip-readability directly on photorealistic outputs, inspired by recent advances in perceptual loss for facial animation \cite{emote,zhuang2024learn2talk3dtalkingface}.
Unlike prior work that computes losses on intermediate mesh representations, our method leverages an efficient \ac{3dgs} pipeline to supervise the final visual output.
This strategy closes the gap between geometric accuracy and perceptual intelligibility.
As illustrated in Figure~\ref{fig:lips}, our \ac{3dgs} renders achieve a visual fidelity far closer to the ground truth than mesh renders, making them a more effective target for perceptual evaluation.

To implement this, we employ the pretrained AutoAVSR model \cite{autoavsr_ma2023} to extract visual speech features, establishing a direct optimization path from the predicted FLAME mesh to the animation's perceptual quality.
For computational efficiency, we render only the 96x96 pixel lip region, which is isolated using a virtual camera positioned via reprojected 3D landmarks.

A potential concern is the domain gap between synthetic \ac{3dgs} renders and real videos.
To validate that our renders serve as a suitable proxy for training, we measured feature similarity within AutoAVSR's embedding space.
We computed the per-frame cosine similarity between features from a ground-truth video and our render of the corresponding ground-truth mesh.
Across our subset of the MEAD dataset, the matching pairs achieve a high mean cosine similarity of $0.697$, while mismatched pairs fall to $0.190$.
These results demonstrate that our 3DGS renders are well-aligned with the AVSR embedding space, validating their suitability as supervision targets.
See supplementary material \ref{sec:supp_cm} for full confusion matrix.

\subsection{Architecture}
\acs{NAME} employs an encoder-decoder transformer architecture, drawing inspiration from Faceformer \cite{faceformer2022}, as illustrated in Figure~\ref{fig:arch}.
The model is designed to generate 3D facial animations from either audio or text input.

\textbf{Encoder.}
The encoder's role is to process the input modality, audio or text, and produce a sequence of feature embeddings aligned with the video frame rate.

During training, or audio-driven inference, the model processes input waveforms using a pretrained Wav2Vec2.0 model \cite{baevski2020wav2vec20frameworkselfsupervised}.
An initial \ac{tcn} extracts low-level features, which are then temporally interpolated to match the 30 FPS video frame rate, ensuring synchronization between the audio and visual streams.
These features are passed through Wav2Vec2.0's transformer encoder to capture long-range contextual dependencies.
To preserve the powerful, generalized audio knowledge, the TCN's weights are frozen, while the subsequent transformer layers are trained end-to-end to adapt them to the facial animation task.

To enable direct text-to-mouthing synthesis, the framework integrates a pretrained F5-TTS model.
During inference, input text is first converted into a synthetic audio waveform by the TTS model. This waveform is then processed by the same Wav2Vec2.0 encoder, creating a seamless pipeline from text to animation.

Regardless of input modality, the encoder's final output is a sequence of feature embeddings, $I_T=(i_1,...,i_T)$, which are the cross-modal input given to our decoder.

\textbf{Decoder.}
The decoder autoregressively predicts vertex offsets that deform a neutral FLAME mesh to create the final animation.
At each step, the model predicts the next vertex offset, $\hat{\textbf{V}}_t$, given the past offsets $\hat{\textbf{V}}_{<t}$, the speaker embedding $s_n$, and the input features $I_T$:
\begin{equation}
    \text{Model}(\hat{\textbf{V}}_{<t}, s_n, I_T) \rightarrow(\hat{\textbf{V}}_t)
    \label{eq:transformer}
\end{equation}

The per-subject neutral mesh serves a dual purpose: it provides the base geometry to which offsets are added, and it acts as an identity prior.
Its vertices are passed through a linear layer to produce a speaker embedding, $s_n$, which conditions the decoder to generate subject-specific morphology and articulation styles by fusing it to the projected past offsets.

During training, the sequence of past vertex offsets is projected into a 64-dimensional embedding space and fused with the speaker embedding and periodic positional encodings.
This combined sequence is processed by a single transformer decoder layer composed of self-attention with a temporal bias, cross-attention with an alignment bias, and a feed-forward network.
The decoder uses four attention heads and a dropout rate of 0.3.
The resulting output is mapped back to the vertex offset space and added to the neutral mesh to produce the final animated mesh, $\textbf{V}_t$.
For stable training, we employ teacher forcing and a fixed learning rate, as in Faceformer \cite{faceformer2022}.

\subsection{Supervision}

We train our model using a three-stage curriculum that strategically combines a standard geometric loss with our novel perceptual supervision.
This approach avoids instability caused by applying the complex perceptual loss from the outset before a reasonable audio-to-geometry mapping is learned.
Furthermore, it manages the comparatively large computational cost of differentiable rendering by introducing it only in the final refinement stage.

Our first stage is geometric pretraining, by using VOCASET \cite{voca}, a dataset with high-quality 3D ground truth.
This stage is supervised only by geometric vertex loss, $\mathcal{L}_{\text{vert}}$, to learn the mapping between input and facial movements.
The second stage is the transition to MEAD \cite{kaisiyuan2020mead} to adapt the model to the changes between the high quality 3D reconstruction and the pseudo-ground truth generated meshes.
We continue to train with only $\mathcal{L}_{\text{vert}}$, but adjust weighting for focus on the lip region and account for noise in the pseudo-GT data.
Finally, we introduce the novel lip-reading loss $\mathcal{L}_{\text{read}}$ and fine-tune the model on MEAD with a combined loss function.

The primary geometric loss term, $\mathcal{L}_{\text{vert}}$, is a standard vertex loss that minimizes the distance between the predicted mesh vertices and the pseudo-ground truth mesh vertices.
It is calculated as a weighted \ac{mse}:
\begin{equation}
    \mathcal{L}_{\text{vert}}=\sum^T_{t=1}\sum^V_{v=1}(||\hat{\textbf{V}}_{t,v}-{\textbf{V}}_{t,v}||^2)W_v,
    \label{eq:mse}
\end{equation}
where $V$ is the total number of vertices, 5143, of the FLAME mesh with added teeth, and $W_v$ is a per-vertex weight.
During the first stage, we set $W_v$ to 1.0 for all vertices.
In the second and third stages, we reduce the weight for all non-skin vertices to 0.5 to reduce the influence of noisy pseudo-ground truth in those areas, and to 0.0 for eye vertices to ignore irrelevant reading motions present in the dataset.

Our novel perceptual loss, $\mathcal{L}_{\text{read}}$, directly evaluates the visual quality of the mouth motion in the rendered image space.
First, we use the differentiable renderer to generate a sequence of lip-region images, $\hat{\textbf{I}}_T$, from the predicted mesh sequence. 
Then we use a pre-trained AutoAVSR \cite{autoavsr_ma2023} to produce lip-reading features for both the predicted sequences, $\hat{\textbf{I}}_T$, and the input frames, ${\textbf{I}}_T$.
The loss is the cosine distance between these feature embeddings:
\begin{equation}
    \mathcal{L}_{\text{read}} = 1 - \text{CosSim}(\text{AutoAVSR}({\textbf{I}}_T), \text{AutoAVSR}(\hat{\textbf{I}}_T)).
    \label{eq:read}
\end{equation}
To reduce computational cost, the features from the ground truth frames are precomputed and stored.

Lip-reading supervision is applied only during the third stage of training.
The total loss function is a weighted sum of the geometric and perceptual losses:
\begin{equation}
    \mathcal{L} = \mathcal{L}_{\text{vert}} + \lambda_{\text{read}}\mathcal{L}_{\text{read}},
    \label{eq:loss}
\end{equation}

where $\lambda_{\text{read}} = 1e-5$.
This value was chosen empirically to scale the magnitude of $\mathcal{L}_{\text{read}}$ to be comparable to that of $\mathcal{L}_{\text{vert}}$ during training, ensuring a balanced contribution from both supervision signals.
Lower values, such as $1e-6$, yielded negligible improvements over the baseline, while higher values caused artifacts, including mesh protrusions and sharp angles.

\section{Experiments}
\begin{figure*}[!h]
    \includegraphics[width=\linewidth]{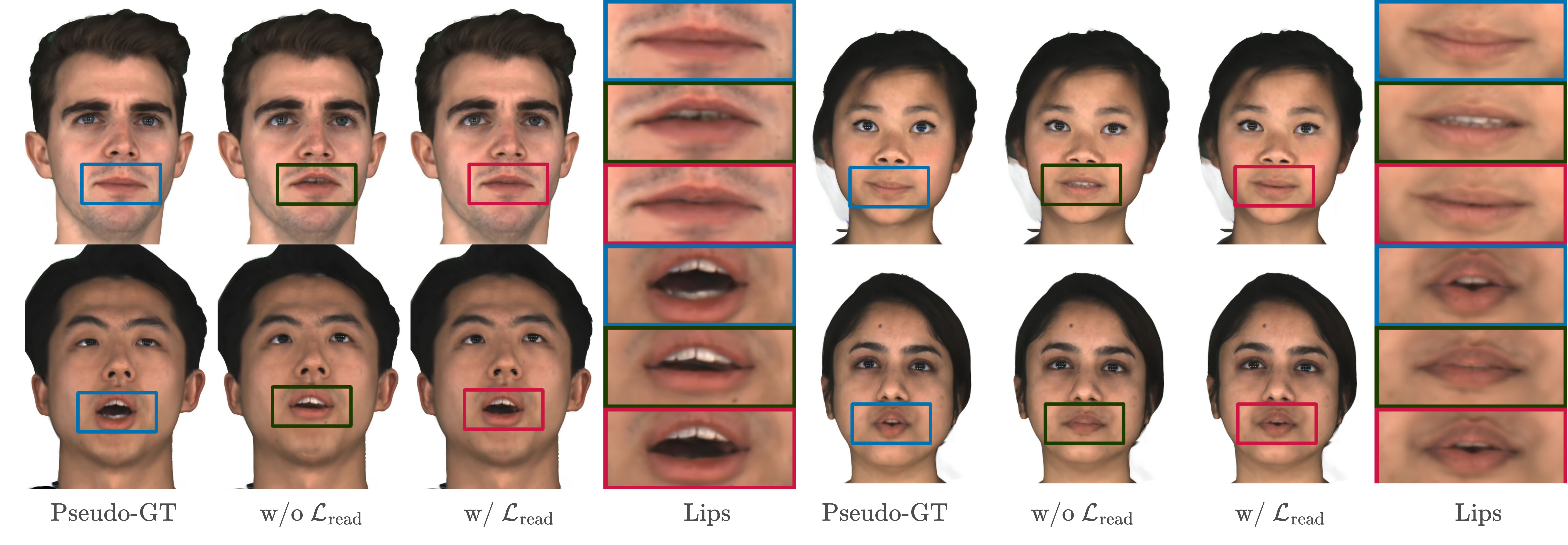}
    \caption{\textbf{Qualitative Results.} Visual comparisons for four unseen subjects and sentences from MEAD \cite{kaisiyuan2020mead}, highlighting how \acs{NAME} better preserves lip articulation than the baseline. Each subfigure displays three frames, left to right: Pseudo-GT render, \acs{NAME} without lip-reading loss ($\mathcal{L}_{\text{read}}$), and \acs{NAME} with full supervision. Next to these, we show a zoomed-in crop of the mouth region in the same order top to bottom, highlighting differences in lip articulation.}
    \label{fig:qualitative}
\end{figure*}

To evaluate the effectiveness of \acs{NAME}, particularly the contribution of the 3DGS-based lip-reading loss, we conducted a series of experiments using the VOCASET \cite{voca} and MEAD \cite{kaisiyuan2020mead} datasets.
Our primary analysis focuses on an ablation study comparing our full model against a variant trained without this perceptual supervision.
The results demonstrate that incorporating a \ac{3dgs}-based lip-reading loss significantly improves lip-motion accuracy and perceived realism over a strong baseline, without degrading overall visual quality.

All models were implemented in PyTorch and trained on an NVIDIA 3090, with a constant learning rate of $1e-4$ and the Adam optimizer \cite{adam}.
The initial stages focusing on mesh supervision were trained for 250 epochs with a batch size of four. The final stage, incorporating the lip-reading supervision, was subsequently trained for 100 epochs using a batch size of one with gradient accumulation over four steps due to memory constraints.

\subsection{Datasets and Preprocessing}
For initial geometric pre-training, we use VOCASET \cite{voca}, which provides 480 sequences of high-quality 3D mesh data and aligned audio for 12 subjects. 
We follow  the standard 8/2/2 subject split for training, validation, and testing, ensuring no overlap of subjects or sentences exists.
Additionally, the provided 60 FPS mesh data is resampled to 30 FPS for consistency.

As VOCASET lacks any video data, we opt to use the multi-view audiovisual dataset MEAD \cite{kaisiyuan2020mead} to create our photorealistic head avatars and test our lip-reading loss.
We use the 40 neutral emotion sequences of the 48 actors provided.

To create the pseudo-ground truth FLAME meshes, we fit the FLAME model to the video data using the methodology outlined in Section \ref{sec:prelim}.
Due to tracking and masking challenges, five subjects were excluded. 
For fair evaluation, we split into six validation and test subjects with both sets containing two female and four male.
As all 40 sentences are spoken by all subjects, we train only on the first 30, leaving the remaining 10 sentences to be used across the validation and test subject sets.

\subsection{Quantitative Results}
To evaluate the geometric accuracy of the lip movements generated by our model, we employ the standard \acf{lve} metric \cite{richard2021meshtalk}.
\ac{lve} computes the maximum per-frame L2 distance between predicted and ground truth lip vertices, averaged across the sequence and is an key metric for the geometric accuracy of the animation.
The average \ac{lve} scores for different stages of our pipeline, evaluated on the test sets of VOCASET and MEAD, are presented in Table~\ref{tab:lve}.

\begin{table}[!h]
    \renewcommand{\arraystretch}{1.1} 
    \setlength{\tabcolsep}{3pt}

    \begin{tabular}{lcc}
    \toprule
    \textbf{Stage / Method}                       & $\downarrow$ \textbf{LVE}$_{\textbf{VOCASET}}$  & $\downarrow$ \textbf{LVE}$_{\textbf{MEAD}}$  \\
    \midrule
    Pretraining                             & 3.06                              & 7.05                            \\
    \acs{NAME} w/o $\mathcal{L}_{\text{read}}$    & --                                & 3.85                             \\
    \textbf{\acs{NAME}}                           & \textbf{--}                       & \textbf{1.69}                    \\
    \bottomrule
    \end{tabular}
    \caption{\textbf{Comparison of Loss Vertex Error} (LVE (mm), lower is better) across pipeline stages. $\mathcal{L}_{\text{read}}$ represents the lip-reading loss.}
    \label{tab:lve}
\end{table}

The inherent differences between the datasets and the challenge posed by MEAD are immediately apparent.
The model pretrained on VOCASET achieves an \ac{lve} of \SI{3.06}{\milli\meter} on its native test set. However, when this same model is applied to the MEAD test set, the \ac{lve} significantly increases to \SI{7.05}{\milli\meter}, a rise of 130\%.
This likely reflects differences in capture conditions, with MEAD’s pseudo-ground truth being noisier or more variable than VOCASET’s direct 3D scans, possibly due to tracking intricacies or less controlled recording conditions.
Adapting to MEAD by fine-tuning mitigates this, improving the \ac{lve} on MEAD to \SI{3.85}{\milli\meter} and yet remaining 26\% higher than the retrained models performance on VOCASET.

Incorporating $\mathcal{L}_{\text{read}}$ in the full \acs{NAME} model further reduces the MEAD \ac{lve} to \SI{1.69}{\milli\meter}, representing a 56.1\% improvement over the fine-tuned model and a 44.8\% reduction relative to the pretrained baseline.
This demonstrates that perceptual supervision from lip-reading not only improves alignment with visual intelligibility but also drives the model toward more precise geometric articulation, likely by enhancing attention to the mouth region during training.

We also assess the visual fidelity of the generated avatars with common visual quality metrics, \ac{psnr} \ac{ssim} \cite{SSIM}, and \ac{lpips} \cite{LPIPS}, on the rendered frames. 
We include: (i) ground truth images, (ii) images rendered from pseudo-ground truth vertices, and (iii) outputs from successive stages of our method.
Since our method does not modify the underlying \ac{3dgs} representation, the pseudo-ground truth renders serve as an upper bound for achievable image quality.
Table~\ref{tab:visual} shows the results of these metrics as averages of per-sequence scores on the MEAD test set.

\begin{table}
    \renewcommand{\arraystretch}{1.1} 
    \setlength{\tabcolsep}{5pt}
    \begin{tabular}{lccc}
    \toprule
    \textbf{Stage / Method}                     & \textbf{PSNR} $\uparrow$  & \textbf{SSIM} $\uparrow$  & \textbf{LPIPS} $\downarrow$ \\
    \midrule
    \textit{Pseudo-GT Vertices}                 & \textit{20.47}            & \textit{0.9126}           & \textit{0.1265}       \\ 
    Pretraining                           & \textbf{19.48}                    & 0.9057                       & 0.1353         \\
    \acs{NAME} w/o $\mathcal{L}_{\text{read}}$  & 19.29                     & 0.9077                      & 0.1326         \\
    \textbf{\acs{NAME}}                         & 19.32             & \textbf{0.9083}             & \textbf{0.1316}  \\
    \bottomrule
    \end{tabular}
    \caption{\textbf{Visual Results.} PSNR, SSIM, and LPIPS scores for different training stages on the MEAD test set.
    }
    \label{tab:visual}
\end{table}

\begin{figure}
    \begin{subfigure}{0.32\linewidth}
        \includegraphics[width=\linewidth]{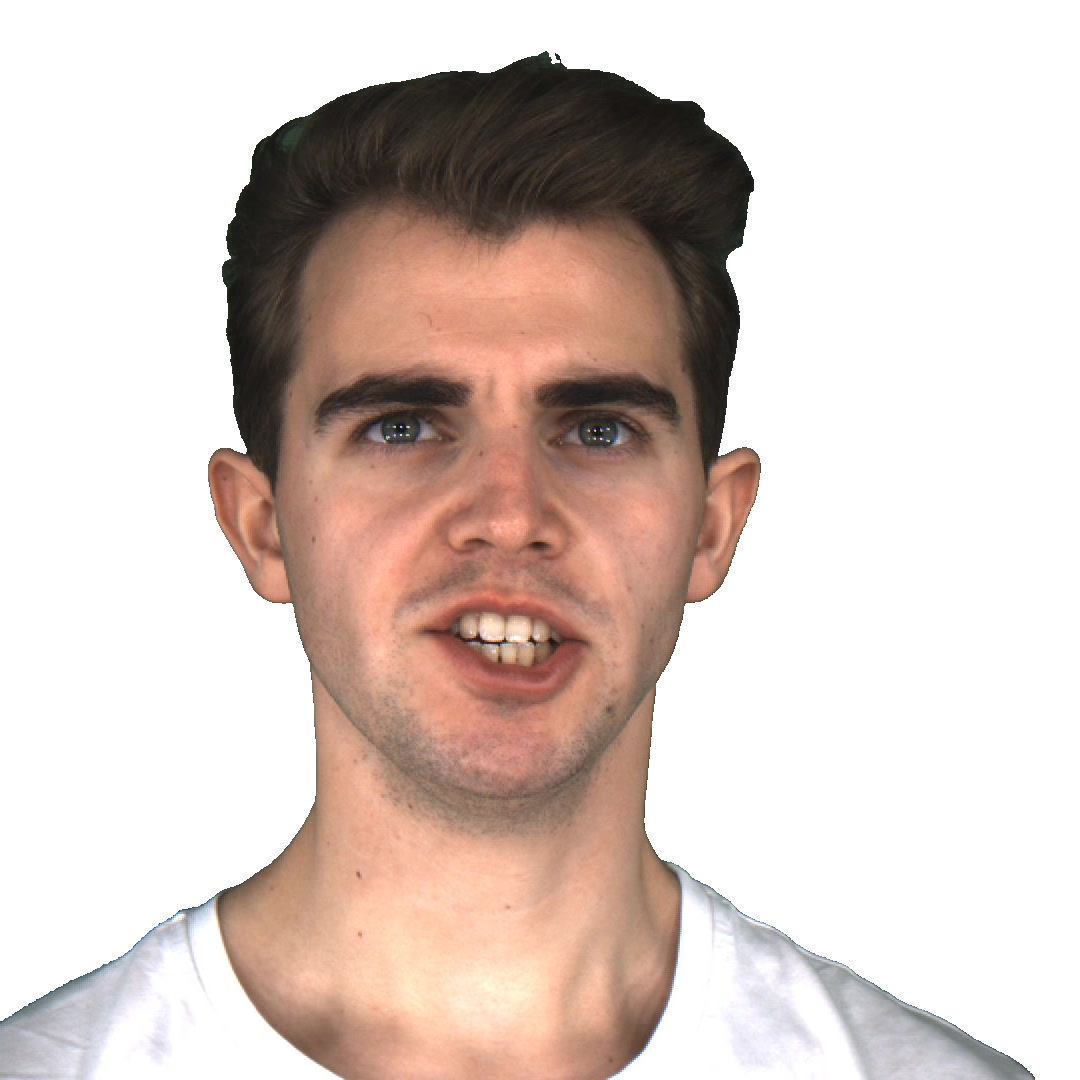}
        \caption{Ground Truth}
    \end{subfigure}
    \hfill
    \begin{subfigure}{0.32\linewidth}
        \includegraphics[width=\linewidth]{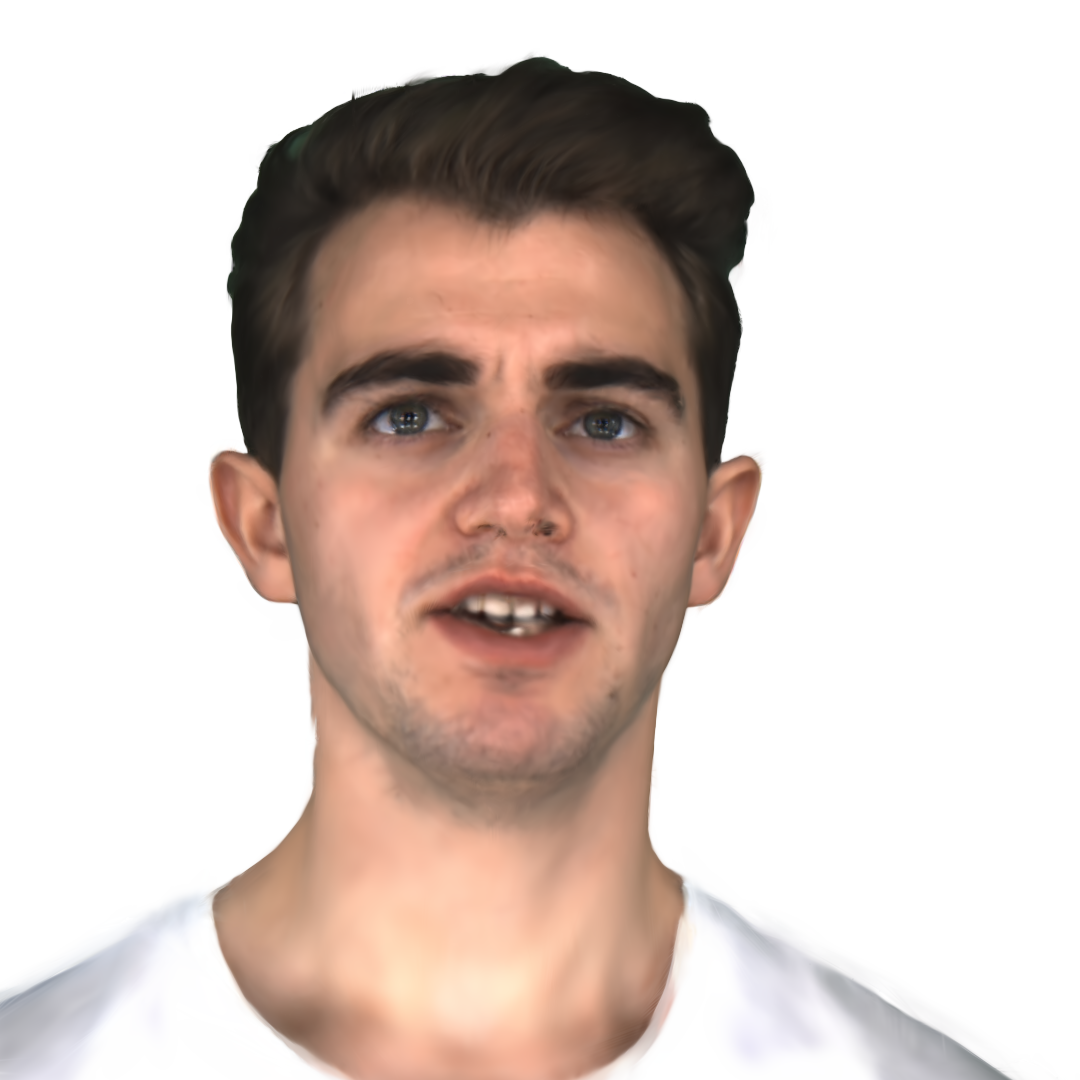}
        \caption{\acs{NAME}}
    \end{subfigure}
    \hfill
    \begin{subfigure}{0.32\linewidth}
        \includegraphics[width=\linewidth]{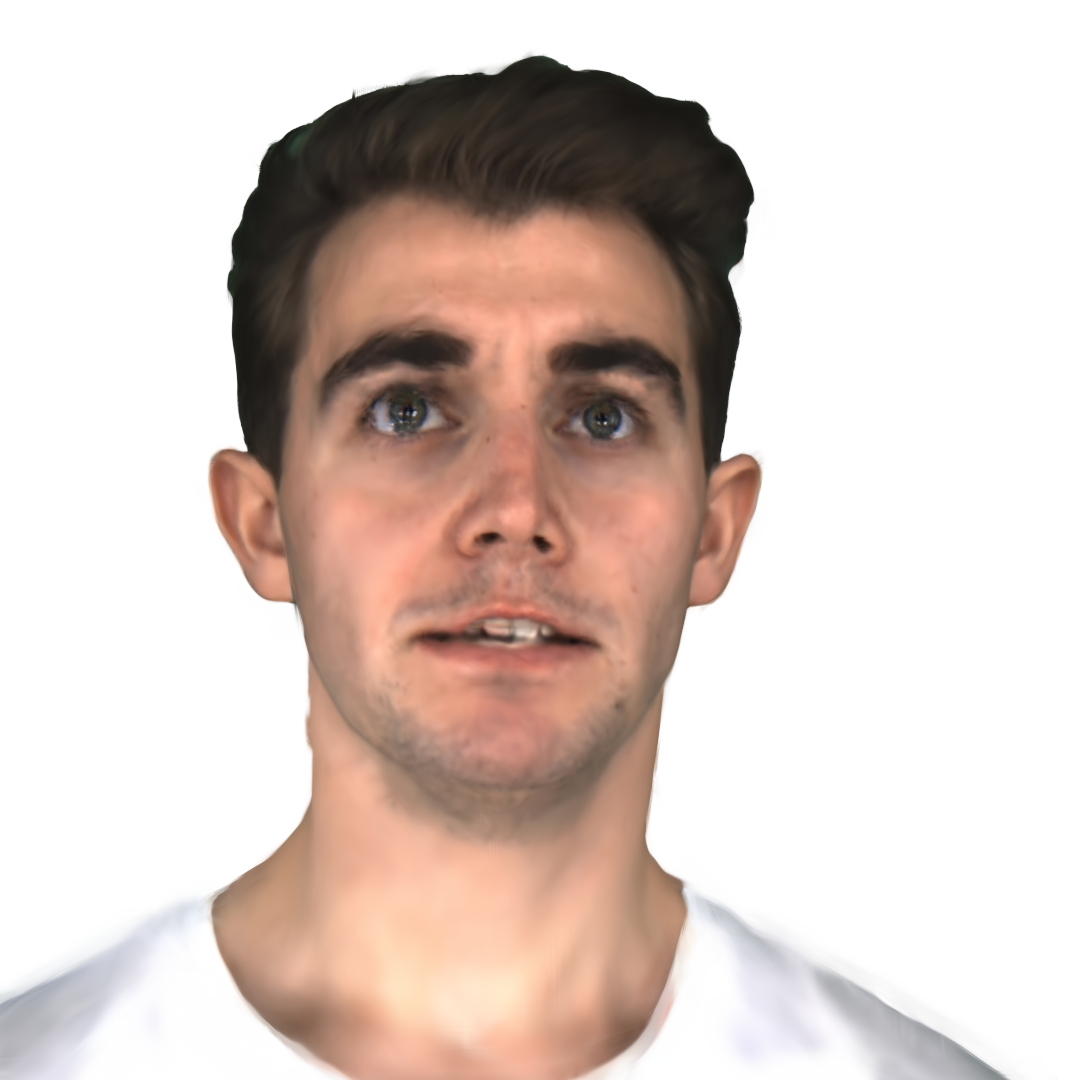}
        \caption{Pretrained}
    \end{subfigure}
    \caption{\textbf{Qualitative Comparison.} Example outputs from \acs{NAME} and the model pretrained only on VOCASET, on unseen subjects and sentences from MEAD \cite{kaisiyuan2020mead}. Despite the clear perceptual differences, the PSNR values for these frames are 21.11\,dB and 20.90\,dB, respectively.}
    \label{fig:quant_comp}
\end{figure}

As shown in Table~\ref{tab:visual}, improvements in these metrics across training stages are modest.
The full \acs{NAME} model yields a slight gain in \ac{ssim} and \ac{lpips}, though \ac{psnr} decreases slightly compared to the pretrained model.
This discrepancy likely arises because these metrics are more sensitive to global image fidelity than to the local articulatory details, such as the lips, that our method explicitly targets.
Figure~\ref{fig:quant_comp} illustrates this, showing that while both the \acs{NAME} and pretraining models receive similar \ac{psnr} scores, the renders from the pretrained model appear visually unnatural to humans, with distorted expressions, while \acs{NAME} produces more realistic facial dynamics.
These results indicate that our pipeline preserves visual quality relative to the pseudo-ground truth ceiling, and that introducing perceptual lip-reading supervision does not degrade image quality, despite not being directly optimized for pixel fidelity.

\subsection{Qualitative Results and User Study}
Quantitative metrics, both geometric and visual, do not fully capture the perceptual quality of dynamic facial animations as experienced by human observers.
Therefore, qualitative analysis and user studies are essential to evaluate the effectiveness of our method in generating realistic and intelligible lip movements.
Figure~\ref{fig:qualitative} presents a series of visual comparisons between our full \acs{NAME} model, the baseline model (\acs{NAME} w/o $\mathcal{L}_{\text{read}}$), and renders from pseudo-ground truth vertices for unseen subjects and sentences from the MEAD test set.
We use pseudo-ground truth vertices driven avatar, rather than real images, as a reference to isolate and evaluate the accuracy of motion synthesis, independent of texture or identity reconstruction.

Visually, the \acs{NAME} model produces animations that surpass the baseline in key aspects.
For example, improved mouth closures are evident in the top row of Figure~\ref{fig:qualitative}, while the bottom right demonstrates more expressive, large-scale lip motions.
The bottom left highlights improvements in generating distinct mouth shapes, such as pursed lips.

To further assess the perceptual quality of our generated animations, we conducted a user study.
We recruited 51 participants, who were presented with side-by-side video comparisons.
Each comparison showed animations generated by \acs{NAME} versus either a baseline model (trained without lip-reading supervision), or animations rendered from pseudo-ground truth meshes.
Participants were instructed to rate their preference based on visual realism of lip movements and ease of lip-reading, using a 5-point scale from \textit{strongly prefer left} (-2), \textit{prefer left} (-1), \textit{no preference} (0), \textit{prefer right} (1), to \textit{strongly prefer right} (2).
Each participant evaluated 20 randomly selected video pairs from a pool of 100 test sequences, ensuring broad coverage while maintaining manageable session length.
The study interface, detailed instructions, and example comparisons interface can be seen in Supplementary~\ref{sec:user_study}.
Preference scores were computed from these ratings to quantify perceptual advantages between methods.

\begin{table}
    \setlength{\tabcolsep}{4pt}
    \renewcommand{\arraystretch}{1.1}
    \begin{tabular}{lcc}
    \toprule
    \textbf{Comparison Pair} & Realism (\%) $\uparrow$ & Lip Clarity (\%) $\uparrow$\\
    \midrule
    Ours vs. Baseline    & $63.8\%\pm8.8\%$ & $66.6\%\pm10.5\%$ \\
    Ours vs. Pseudo-GT   & $34.9\%\pm8.0\%$ & $33.6\%\pm8.7\%$ \\
    \bottomrule
    \end{tabular}
    \caption{\textbf{User Study: Overall Preference.} Percentage of times \acs{NAME} was preferred in A/B comparisons, $\pm$ the standard deviation. `Baseline' is \acs{NAME} w/o $\mathcal{L}_{\text{read}}$; `Pseudo-GT' uses fitted ground truth vertices.}
    \label{tab:user}
\end{table}

The results of the user study, shown in Table~\ref{tab:user}, indicate that \acs{NAME} outperforms the baseline with 65\% of participants preferring the animations generated by \acs{NAME} in terms of both realism and lip clarity.
This highlights the contribution of lip-reading loss. 

However, when compared to the pseudo-ground-truth vertices, our output quality still has room for improvement. This is likely due to limitations in expressiveness and a relative lack of subtle, fast lip movements in our current generation.
While \acs{NAME} improves lip articulation overall, the model can under-articulate very rapid or subtle consonant closures, such as plosives (/p/, /b/) or brief tongue contacts, which the FLAME model does not explicitly capture. 
Additionally, users reported that eye motion, blinking, and general upper-face activity in the pseudo-ground-truth renders contributed significantly to perceived realism.
This suggests that the lower gap rating for our method may be partly due to missing or under-articulated non-verbal cues beyond the mouth region, which we currently do not model.
This highlights that while our method advances lip-synchronization, achieving full human-level realism in 3D avatars is a holistic challenge.
Additional work to ingrate our method with systems that control upper-face expressions, eye gaze, and blinking to bridge this remaining gap is a clear direction for future work.

\subsection{Sign Language Production}
\begin{figure}
    \begin{subfigure}[b]{0.1\textwidth}
        \includegraphics[width=\textwidth]{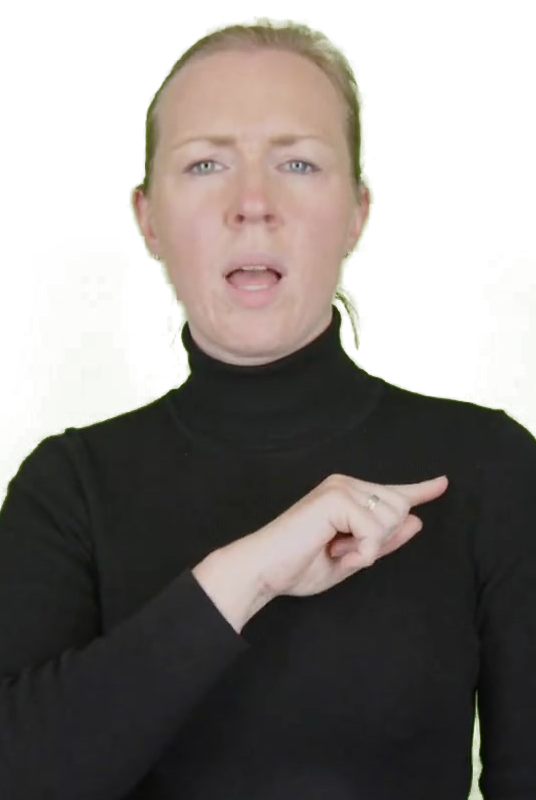}
        \caption{\textbf{Why}}
        \label{fig:slp_why_gt}
    \end{subfigure}
    \hfill
    \begin{subfigure}[b]{0.1\textwidth}
        \includegraphics[width=\textwidth]{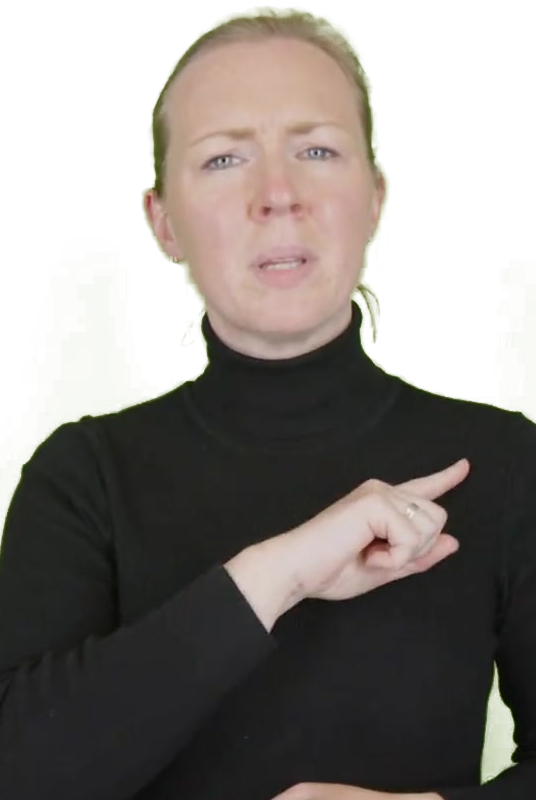}
        \caption{\textbf{Because}}
        \label{fig:slp_bc_gt}
    \end{subfigure}
    \hfill
    \begin{subfigure}[b]{0.1\textwidth}
        \includegraphics[width=\textwidth]{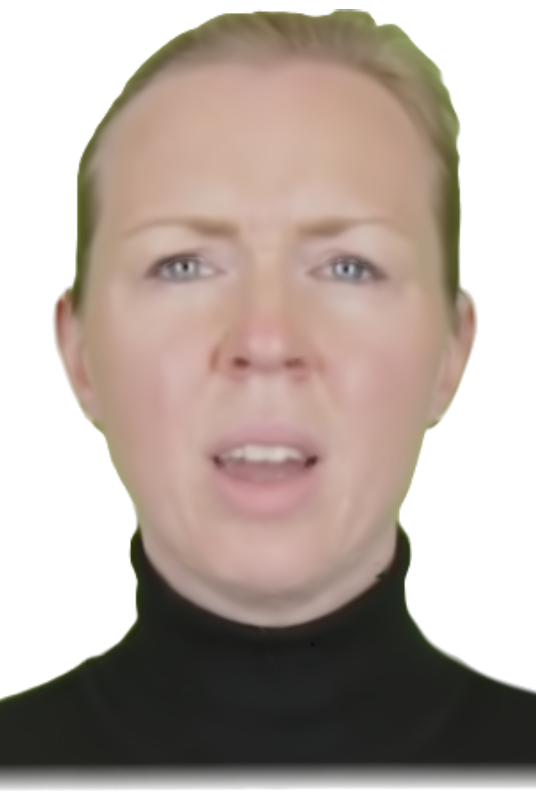}
        \caption{\textit{Why}}
        \label{fig:slp_why}
    \end{subfigure}
    \hfill
    \begin{subfigure}[b]{0.1\textwidth}
        \includegraphics[width=\textwidth]{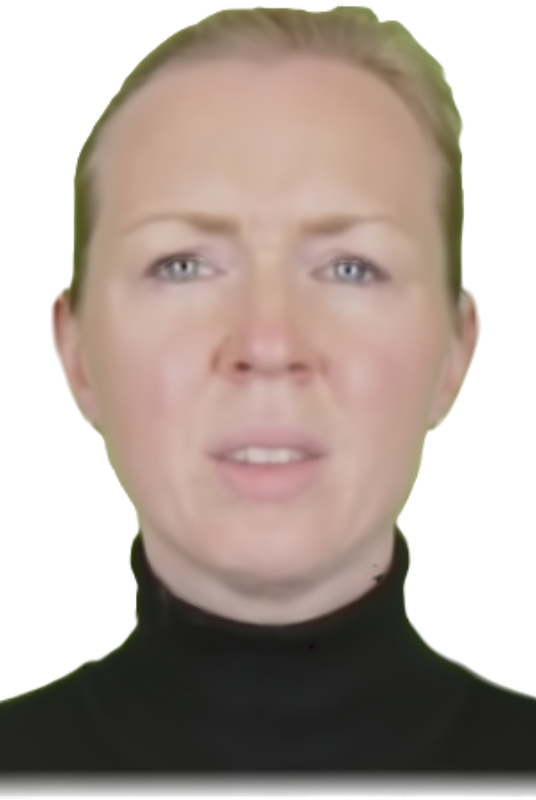}
        \caption{\textit{Because}}
        \label{fig:slp_bc}
    \end{subfigure}
    \caption{\ac{slp} Example. Two left sub-figures show still frames of two \acs{bsl} signs with identical manual and different mouthings. Right sub-figures show how \acs{NAME} can generate mouthings capable of disambiguating signs using text input.}
    \label{fig:slp}
\end{figure}

To evaluate our method's ability to generate linguistically meaningful mouthings, we chose sign pairs which share the same manual component in \ac{bsl}, differing only in their mouthings.
Figure~\ref{fig:slp} shows an example of the `why' and `because' signs.
Still frames of a real, unseen, signer are on the left and outputs from \acs{NAME} generated purely from gloss-level text input and synthesized speech are on the right.
Our model successfully produces distinct mouth shapes for these two signs, capturing differences such as lip rounding and closure patterns. 
This is achieved using only glosses and the \ac{tts} model, without requiring paired audio or manual alignment.
The success of this application is a direct result of the improved lip readability provided by our perceptual loss, demonstrating that optimizing for visual intelligibility enables crucial downstream tasks in accessibility and human-computer interaction that are unattainable with purely geometric supervision.

\section{Conclusion}
We present \acs{NAME}, a method for speech-driven 3D facial animation that bridges geometric accuracy and perceptual intelligibility.
By supervising directly in the rendered domain with a lip-reading loss on photorealistic differentiable 3DGS avatars, our approach achieves a 56\% LVE reduction on MEAD without degrading image fidelity.
A user study confirms clear gains in realism and lip clarity, and we demonstrate practical impact for text-to-mouthing in sign language, where viseme precision is essential.

On the other hand, several limitations remain.
Differentiable 3DGS training incurs high computational cost and limits batch size, suggesting a need for more efficient renderers.
The method also relies on per-subject multi-view avatars; future work could adapt our loss to more generalizable or few-shot avatar pipelines. Finally, our model lacks explicit control of emotion and upper-face cues, which perceptually matter for natural communication.

Overall, our results demonstrate that incorporating direct perceptual supervision at the final output level is a promising step toward more expressive and intelligible 3D avatars for accessible communication in applications such as sign language translation and human-computer interaction.

\section*{Acknowledgements}
This work was supported by the SNSF project `SMILE II' (CRSII5 193686), the Innosuisse IICT Flagship (PFFS-21-47), EPSRC grant APP24554 (SignGPT-EP/Z535370/1) and through funding from Google.org via the AI for Global Goals scheme.
This work reflects only the author's views and the funders are not responsible for any use that may be made of the information it contains.

{
    \small
    \bibliographystyle{ieeenat_fullname}
    \bibliography{main}
}
\clearpage
\setcounter{page}{1}
\maketitlesupplementary

\section{Supplementary Material}
\label{sec:supp}
\subsection{AutoAVSR Feature Alignment}

\begin{figure}[!h]
    \includegraphics[width=.95\linewidth]{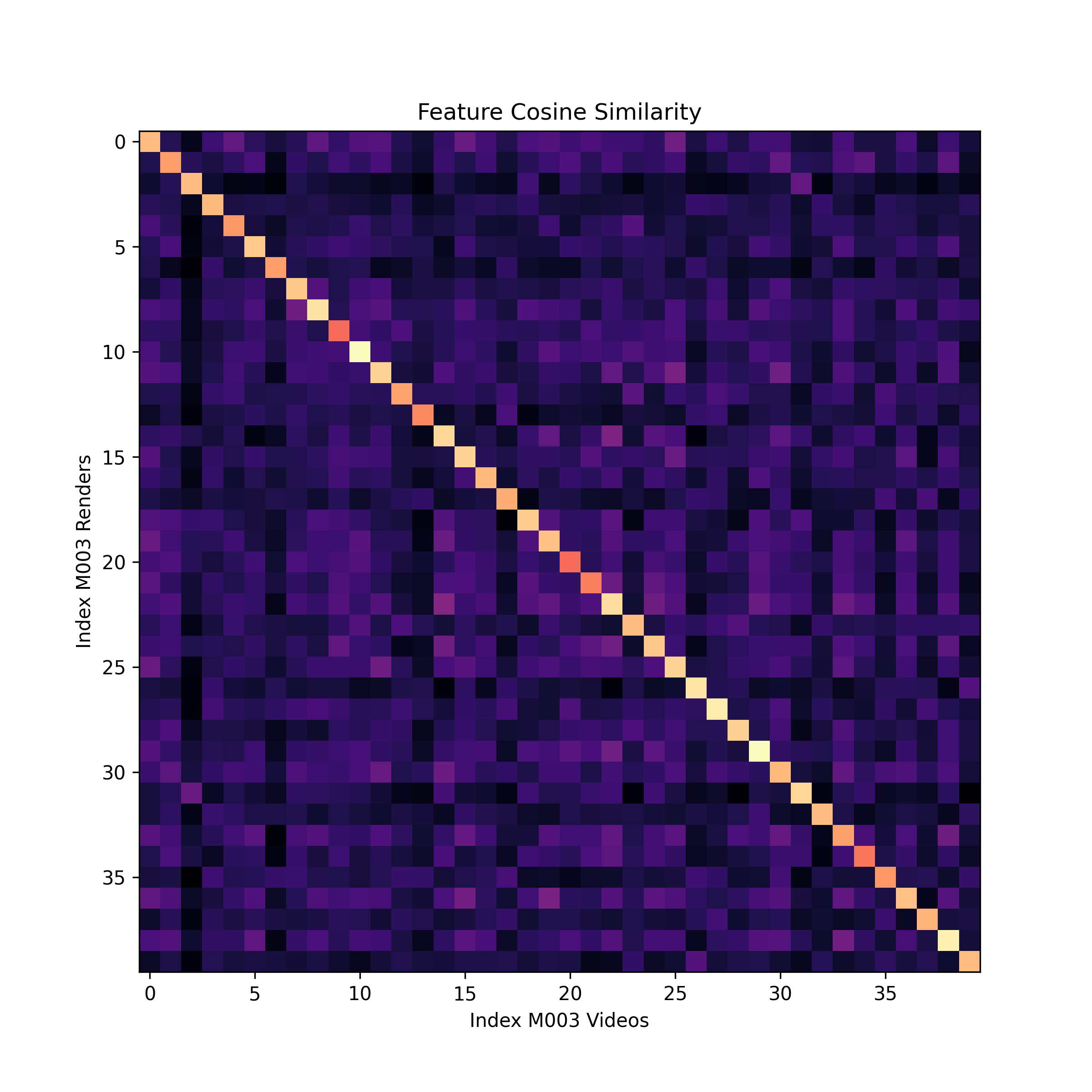}
    \caption{\textbf{Cosine similarity confusion matrix}.}
    \label{fig:supp_cm}
\end{figure}

The confusion matrix in Figure~\ref{fig:supp_cm} shows how cosine similarity scores are strongest along the diagonal, with non-matching videos elsewhere scoring far lower.
This indicates that the features of the \ac{3dgs} render closely match those of the input frames.

\label{sec:supp_cm}

\subsection{User Study Details}
\label{sec:user_study}

\begin{figure}[!h]
    \includegraphics[width=.95\linewidth]{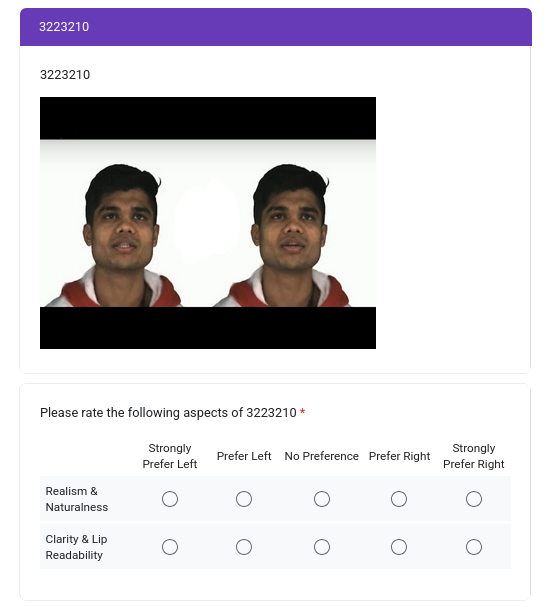}
    \caption{\textbf{Sample User Study Interface}.}
    \label{fig:user_study_sample}
\end{figure}

The user study consisted of 51 participants, each evaluating 20 pairs of videos.
They were recruited via a departmental mailing list.
We circulated 5 variants of the study, each with a different set of 20 pairs, and each video was rated by at least 4 participants.
The participants were asked to rate which video they preferred based on the realism of the lip movements, with the options shown in Figure~\ref{fig:user_study_sample}.
The following instructions were given to the participants:
\begin{quote}
    In this task, you will be presented with pairs of short video animations, shown side-by-side. Each video features an animated 3D character speaking a short sentence.

    Each form contains 20 randomized samples. Your goal is to carefully evaluate and compare them based on two main criteria:
    \begin{itemize}
        \item \textbf{Realism and Naturalness}: How believable, human-like, and natural the lip movements appear.
        \item \textbf{Clarity and Lip Readability}: How clear the lip movements are in representing the spoken words, and how easy it would be to understand what is being said by \textbf{only} watching the lips.
    \end{itemize}
    Audio is provided with the videos. If possible, we kindly request that you use headphones for this task to ensure you can hear clearly.
    
    Please take your time to consider each pair carefully. There are no right or wrong answers.
\end{quote}
The rankings were then converted to a $\{-2,-1,0,1,2\}$ to calculate the preference percentages.

\end{document}